\documentclass{article} % For LaTeX2e
\usepackage{nips13submit_e,times}

\usepackage{graphicx}
\usepackage{latexsym}
\usepackage{amsfonts,amsmath,amssymb}
\usepackage{url}
\usepackage[utf8]{inputenc}
\usepackage{rotating}
\usepackage{fancyref}
\usepackage{hyperref}
\hypersetup{colorlinks=false,pdfborder={0 0 0},}

\title{Unit Tests for Stochastic Optimization}

\author{Tom Schaul%\\ DeepMind Technologies%\\ \texttt{tom@deepmind.com} 
\And Ioannis Antonoglou\\ 
\\ 
DeepMind Technologies\\ 
130 Fenchurch Street, 
London, UK \\
\texttt{\{tom,ioannis,david\}@deepmind.com} 
\And David Silver%\\DeepMind Technologies%\\ \texttt{david@deepmind.com}
}

\nipsfinalcopy

\begin{document}

\maketitle

\begin{abstract}
Optimization by stochastic gradient descent is an important component of
many large-scale machine learning algorithms. A wide variety of such
optimization algorithms have been devised; however, it is unclear
whether these algorithms are robust and widely applicable across many
different optimization landscapes. In this paper we develop a collection
of \emph{unit tests} for stochastic optimization. Each unit test rapidly
evaluates an optimization algorithm on a small-scale, isolated, and
well-understood difficulty, rather than in real-world scenarios where
many such issues are entangled. Passing these unit tests is not
sufficient, but absolutely necessary for any algorithms with claims to
generality or robustness. We give initial quantitative and qualitative
results on numerous established algorithms. The testing framework is
open-source, extensible, and easy to apply to new algorithms.

\end{abstract}

\section{Introduction}

% SGD
Stochastic optimization~\cite{robbins-monro-51} is among the most widely used components in large-scale machine learning, thanks to its linear complexity, efficient data usage, and often superior generalization~\cite{bottou-98x,bottou-lecun-04,bottou-bousquet-08}.
In this context,
numerous variants of stochastic gradient descent have been proposed, in order to improve performance, robustness, or reduce tuning effort~\cite{benveniste-90,leroux-nips-08,bordes-jmlr-09,xu-10,Schaul2013pesky}.
These algorithms may derive from simplifying assumptions on the optimization landscape~\cite{DuchiHS11}, but in practice, they tend to be used as general-purpose tools,
often outside of the space of assumptions their designers intended.
The troublesome conclusion is that practitioners find it difficult to discern where potential
weaknesses of new (or old) algorithms may lie~\cite{pascanu2012understanding}, and when they are applicable -- an issue that is separate from 
raw performance. 
This results in essentially a trial-and-error procedure for finding the appropriate algorithm variant and hyper-parameter settings, every time that the dataset, loss function, regularization parameters, or model architecture change~\cite{glorot10}. 

% unittesting
The objective of this paper is to establish a collection of benchmarks to evaluate stochastic optimization algorithms and guide algorithm design toward robust variants.
Our approach is akin to unit testing, in that it evaluates algorithms on a very broad range of small-scale, isolated, and well-understood difficulties, rather than in real-world scenarios where many such issues are entangled. Passing these unit tests is not sufficient, but absolutely necessary for any algorithms with claims to generality or robustness. 
This is a similar approach to the very fruitful one taken by the black-box optimization community~\cite{hansen2010real, hansen2010comparing}.

% locality assumption
The core assumption we make is that stochastic optimization algorithms are acting \textit{locally}, that is, they
aim for a short-term reduction in loss given the current noisy gradient information, and possibly some internal 
variables that capture local properties of the optimization landscape. These local actions include both approaching nearby optima, and navigating 
slopes, valleys or plateaus that are far from an optimum.
The locality property stems from computational efficiency concerns, but it has the additional benefits of 
minimizing initialization bias and allowing for non-stationary optimization, because properties of the obervation surface observed earlier in the process 
(and their conseuences for the algorithm state) are quickly forgotten.
We therefore concentrate on building local unit tests, that investigate algorithm dynamics on a broad range of local scenarios,
because we expect that detecting local failure modes 
will flag an algorithm as unlikely to be robust on more complex tasks -- and as a first approximation, optimization on such a complex task can
be seen as a sequence of many smaller optimization problems (many of which will not have local optima).

Our divide-and-conquer approach consists of disentangling potential difficulties and testing them in isolation
or in simple couplings. Given that our unit tests are small and quick to evaluate, we can have a much larger 
collection of them, testing hundreds of qualitatively different aspects in less time than it would take to 
optimize a single traditional benchmark to convergence, thus allowing us to spot and address potential weaknesses early.

% details, rest of paper overview
Our main contribution is a testing framework, with unit tests designed to test aspects such as: discontinuous or non-differentiable surfaces, curvature scales, various noise conditions and outliers, saddle-points and plateaus, cliffs and asymmetry, and curl and bootstrapping. It also allows test cases to be concatenated by chaining them in a temporal series, or by combining them into multi-dimensional unit tests (with or without variable coupling).
We give initial quantitative and qualitative results on a number of established algorithms. 

% outlook
We do not expect this to replace traditional benchmark domains that are closer to the real-world, but to complement it in terms of breadth and robustness. We have tried to keep the framework general and extendable, in the hope it will further grow in diversity, and help others in doing robust algorithm design.

\begin{figure}[tb]
\begin{center}
\includegraphics[width=0.98\columnwidth]{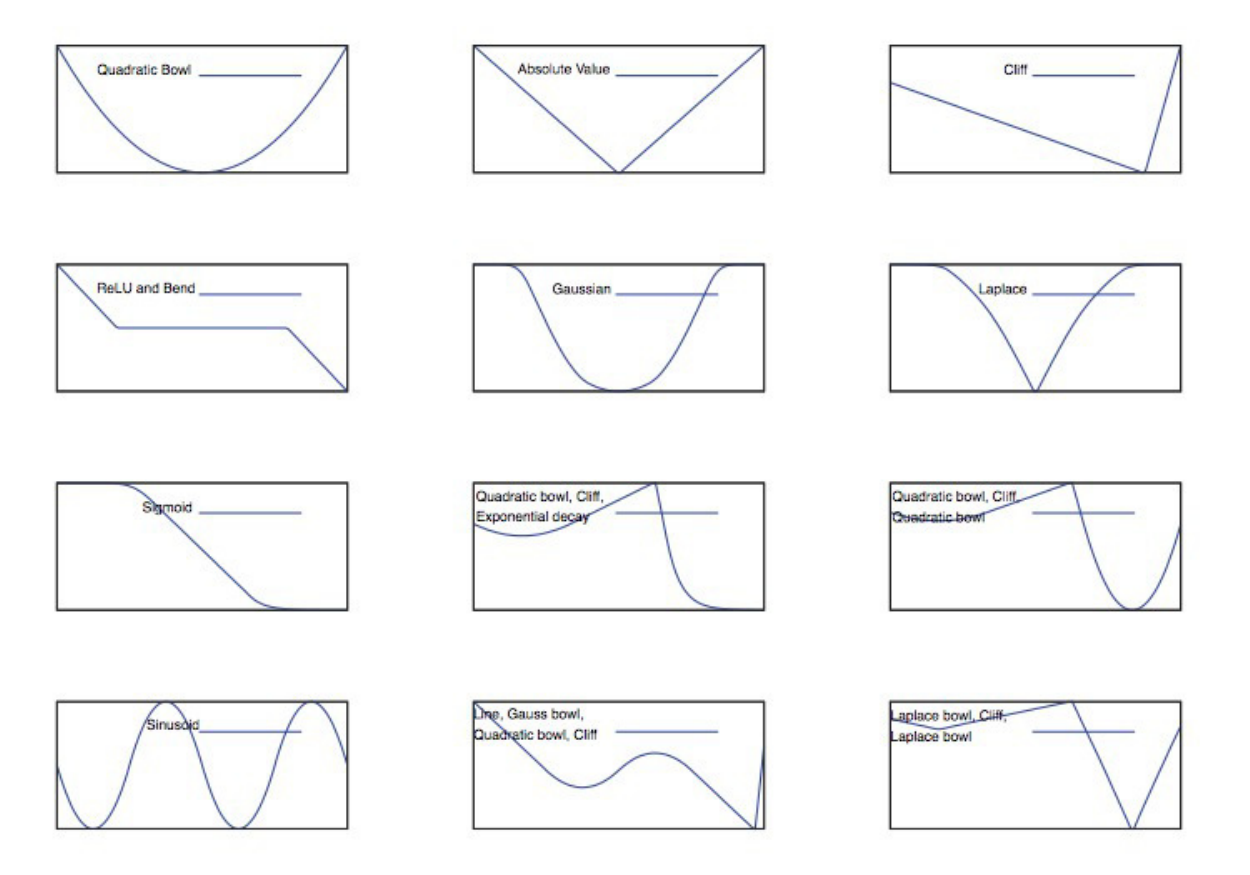}
\caption{\label{Shapefig}
Some one-dimensional shape prototypes. The first six example shapes are atomic prototypes:
a quadratic bowl, 
an absolute value, 
a cliff with a non differential point after which the derivative increases by a factor ten, 
a rectified linear shape followed by a bend, 
an inverse Gaussian, 
an inverse Laplacian.
The next six example shapes are concatenations of atomic prototypes: 
a sigmoid as a concatenation of a non convex Gaussian a line and an exponential, 
a quadratic bowl followed by a cliff and then by an exponential function, 
a quadratic bowl followed by a cliff and another quadratic bowl, 
a sinusoid as a concatenation of quadratic bowls, 
a line followed by a Gaussian bowl, 
a quadratic bowl and a cliff and finally, 
a Laplace bowl followed by a cliff and another Laplace bowl.
}
\end{center}
\end{figure}

\section{Unit test Construction}
Our testing framework is an open-source library containing a collection of unit tests and visualization tools.
Each \emph{unit test} is defined by a prototype function to be optimized, a {prototypical scale}, a {noise prototype}, and optionally a {non-stationarity prototype}. 
A \emph{prototype function} is the concatenation of one or more local {shape prototypes}.
A {multi-dimensional unit test} is a composition of one-dimensional unit tests, optionally with a {rotation prototype} or {curl prototype}.

\subsection{Shape Prototypes}
Shape prototypes are functions defined on an interval, and our collection includes
linear slopes (zero curvature), quadratic curves (fixed curvature), convex or concave curves (varying curvature), and curves with exponentially increasing or decreasing slope.
Further, there are a number of non-differentiable local shape prototypes (absolute value, rectified-linear, cliff).
All of these occur in realistic learning scenarios, for example in logistic regression the loss surface is part concave and part convex, an MSE loss is the prototypical quadratic bowl, but then regularization such as L1 introduces non-differentiable bends (as do rectified-linear or maxout units in deep learning~\cite{krizhevsky2012imagenet,goodfellow2013maxout}). 
Steep cliffs in the loss surface are a common occurrence when training recurrent neural networks, as discussed in~\cite{pascanu2012understanding}.
See the top rows of Figure~\ref{Shapefig} for some examples of shape prototypes.

\subsection{One-dimensional Concatenation}
In our framework, we can chain together a number of shape prototypes, in such a way that the resulting function is continuous and differentiable at all junction points. We can thus produce many prototype functions that closely mimic existing functions, e.g., the Laplace function, sinusoids, saddle-points, step-functions, etc.
See the bottom rows of Figure~\ref{Shapefig} for some examples.

A single \emph{scale} parameter determines the scaling of a concatenated function across all its shapes using the junction constraints. Varying the scales is an important aspect of testing robustness because it is not possible to guarantee well-scaled gradients without substantial overhead. In many learning problems, effort is put into proper normalization~\cite{lecun-98b}, but that is insufficient to guarantee homogeneous scaling, for example throughout all the layers of a deep neural network.

\begin{figure}[tb]
\begin{center}
\includegraphics[width=0.7\columnwidth]{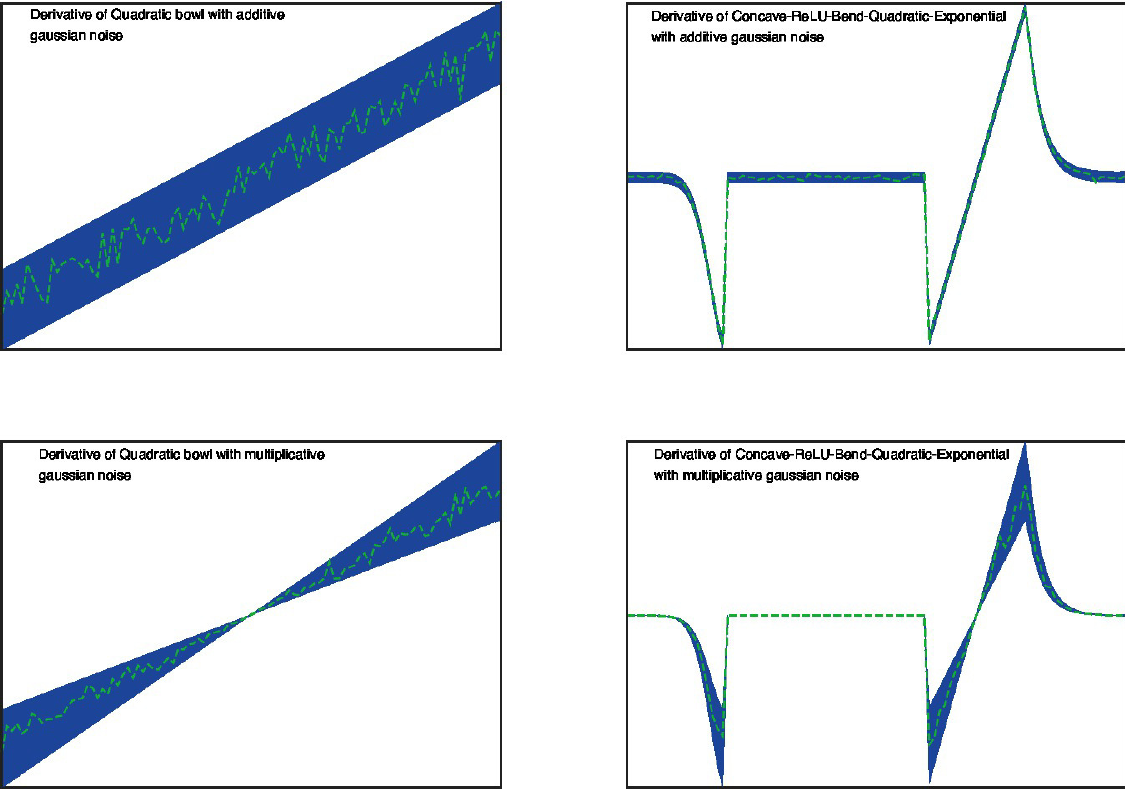}
\caption{\label{Noisefig} Examples of noise applied on prototype functions, green dashed are typical sample gradients, and the standard deviation range is the blue area. The upper two subplots depict Gaussian additive noise, while the lower two show Gaussian multiplicative noise. In the left column, the noise is applied to the gradients of a quadratic bowl prototype (note how the multiplicative noise goes to zero around the optimum in the middle), and on the right it is applied to a concatenation of prototypes.}
\end{center}
\end{figure}

\subsection{Noise Prototypes}

The distinguishing feature of stochastic gradient optimization (compared to batch methods) is that it relies on sample gradients (coming from a subset of even a single element of the dataset) which are inherently noisy. In out unit tests, we model this by four types of stochasticity:
\begin{itemize}
\item Scale-independent additive Gaussian noise on the gradients, which is equivalent to random translations of inputs in a linear model with MSE loss. Note that this type of noise flips the sign of the gradient near the optimum and makes it difficult to approach precisely.
\item Multiplicative (scale-dependent) Gaussian noise on the gradients, which multiplies the gradients by a positive random number (signs are preserved). This corresponds to a learning scenario where the loss curvature is different for different samples near the current point. 
\item Additive zero-median Cauchy noise, mimicking the presence of outliers in the dataset.
\item Mask-out noise, which zeros the gradient (independently for each dimension) with a certain probability. This mimics both training with drop-out~\cite{hinton2012improving}, and scenarios with rectified linear units where a unit will be inactive for some input samples, but not for others. 
\end{itemize}
For the first three, we can vary the noise scale, while for mask-out we pick a drop-out frequency. 
This noise is not necessarily unbiased (as in the Cauchy case), breaking common assumptions made in algorithm design 
(but the modifications in section~\ref{sec:curl} are even worse).
See Figure~\ref{Noisefig} for an illustration of the first two noise prototypes. 
Noise prototypes and prototype functions can be combined independently into one-dimensional unit tests.

\begin{figure}[tb]
\begin{center}
\includegraphics[width=0.9\columnwidth]{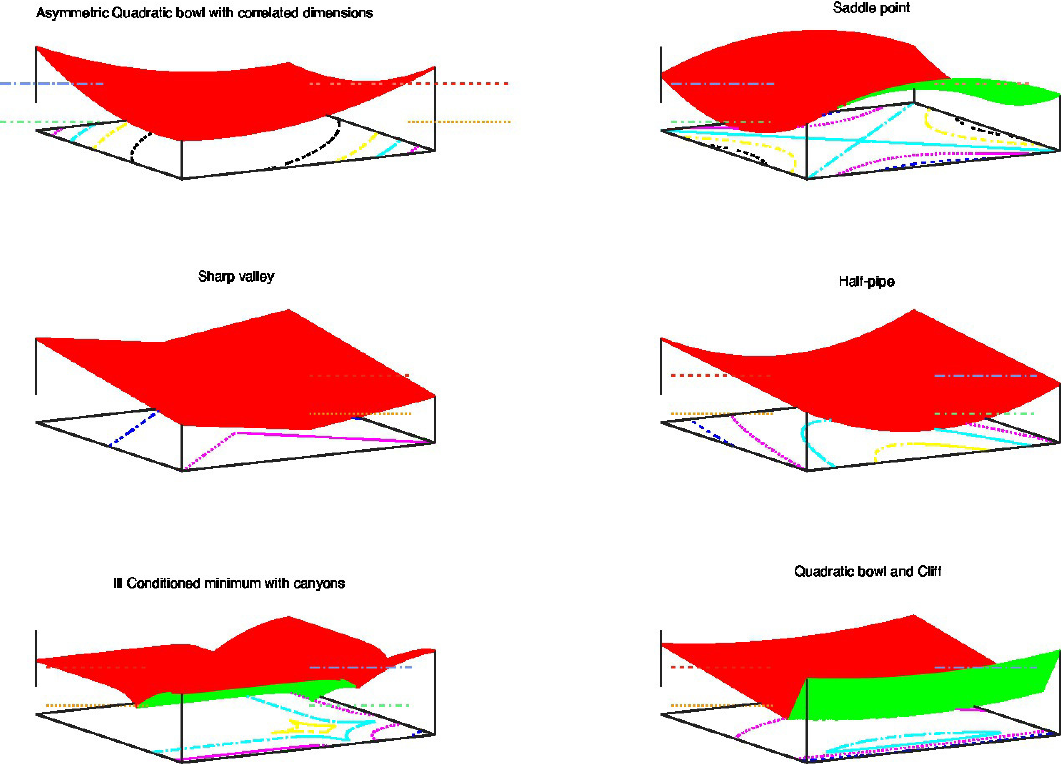}
\caption{Examples of multivariate prototypes. The first subplot depicts an asymmetric quadratic bowl with correlated dimensions, the second a surface with a saddle point, the third a sharp valley surface, the fourth a half-pipe surface where the first dimension is a line and the second one a quadratic bowl. The fifth subplot depicts a surface with an ill conditioned minimum in the point where the two canyons overlap. The surface in the last subplot is the composition of a quadratic bowl in the first dimension and of a cliff in the second.}
\end{center}
\end{figure}

\subsection{Multi-dimensional Composition}
A whole range of difficulties for optimization only exist in higher dimensional parameter spaces (e.g., saddle points, conditoning, correlation).
Therefore, we build high-dimensional unit tests by composing together one-dimensional unit tests. 
For example for two one-dimensional prototype shapes $\mathcal{L}_a$ and $\mathcal{L}_b$ combined with a $p$-norm, the composition is
$\mathcal{L}_{(a,b)}(\theta) = \left( \mathcal{L}_a(\theta_1)^p + \mathcal{L}_b(\theta_2)^p  \right)^{\frac{1}{p}}$.
Noise prototypes are composed independently of shape prototypes.
While they may be composed of concatenated one-dimensional prototypes, higher-dimensional prototypes are not concatenated themselves.
Various levels of \textit{conditioning} can be achieved by having dramatically different scales in different component dimensions. 
%The current test suite contains prototypes in dimensions 2 and 10.

In addition to the choice of prototypes to be combined, and their scale, we permit a rotation in input space, which couples the dimensions together and avoids axis-alignment. These rotations are particularly important for testing diagonal/element-wise optimization algorithms.

\begin{figure}[tb]
\begin{center}
\includegraphics[width=0.5\columnwidth]{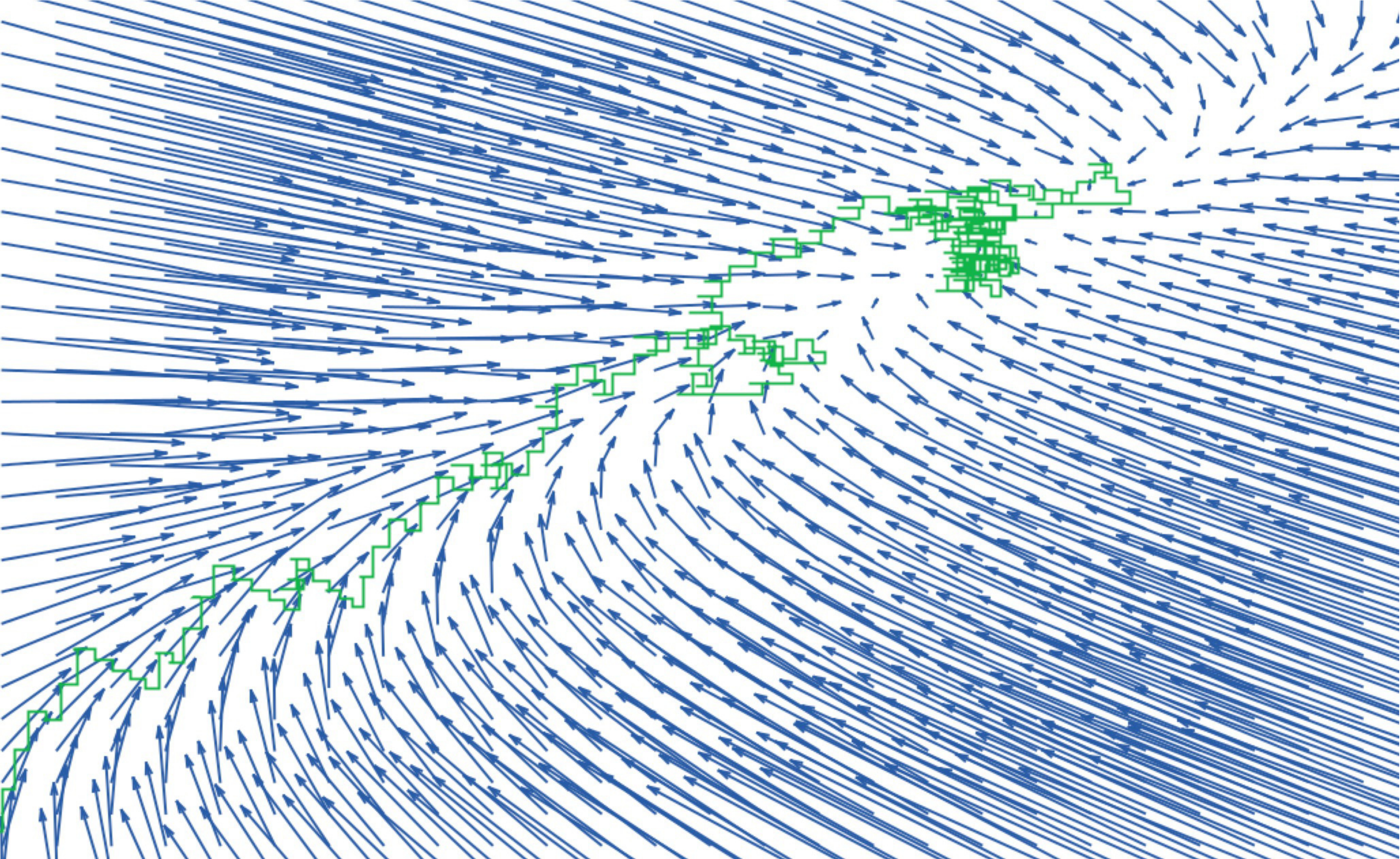}
\caption{\label{Curlfig}
Here, we consider a very simple Markov process, with two states and stochastic transitions between them, and a reward of 0 in the first and of 1 in the second state. 
Consider the parameters of our optimization $\theta$ to be the two state values. Each TD update changes one of them, depending on the stochastic transition observed. In this figure, we plot the vector field of expected update directions (blue arrows) as a function of $\theta$, as well as one sampled trajectory of the TD algorithm. Note how this vector field is not actually a gradient field, but instead has substantial curl, making it a challenging stochastic optimization task.}
\end{center}
\end{figure}

\subsection{Curl}
\label{sec:curl}
In reinforcement learning a value function (the expected discounted reward for each state) can be learned using temporal-difference learning (TD), an update procedure that uses \emph{bootstrapping}: i.e. it pulls the value of the current state towards the value of its successor state \cite{Sutton_Barto_1998}. These stochastic update directions are not proper gradients of any scalar energy field~\cite{Barnard_1993}, but they still form a (more general) vector field with non-zero curl, where the objective for the optimization algorithm is to converge to its fixed-point(s). 
See Figure~\ref{Curlfig} for a detailed example.
We implemented this aspect by allowing different amounts of curl to be added on top of a multi-dimensional vector field in our unit tests, which is done by rotating the produced gradient vectors using a fixed rotation matrix. This is reasonably realistic; in fact, for the TD example in Figure~\ref{Curlfig}, the resulting vector field is exactly the gradient field of a quadratic combined with a (small-angle) rotation.

\subsection{Non-stationarity}
In many settings it is necessary to optimize a non-stationary objective function. This may typically occur in a non-stationary task where the problem to be solved changes over time. However, non-stationary optimization can even be important in large stationary tasks (with temporal structure in the samples), when the algorithm chooses to \emph{track} a particular dynamic aspect of the problem, rather than attempting to converge to a global but static solution of the problem~\cite{Sutton_Koop_Silver_2007}. In addition, reinforcement learning (RL) tasks often involve non-stationary optimization. For example, many RL algorithms proceed by evaluating the value function using the TD algorithm described in the previous section. This results in two sources of non-stationarity: the target value changes at every step (resulting in the previously described curl); and also the state distribution changes as the value function improves and better actions are selected. These scenarios can be therefore be viewed as non-stationary loss functions, but whose optimum moves as a function of the current parameter values.

We test non-stationarity in three different ways. We let the location of the optimum move smoothly, via random translations of the parameter space, or we let the the scale of the shape prototype vary randomly (on average by 10\% in each direction), or, on noisy unit tests, we let the scale of the noise vary randomly. Currently, these changes happen once every 10 steps.
A type of non-stationarity that involves more abrupt switching is discussed in section~\ref{algostates}.

\section{Experiments}

\subsection{Setup and Algorithms}

For our experiments, we test the candidate algorithms on over 3000 unit tests, with up to 10 parameter dimensions. Each algorithm-unit test pairing is repeated 10 times, but with reusing the same 10 random seeds across all algorithms and setups.
For each run $k$, we compute the true expected loss at the parameter value reached after 100 update steps 
$\mathcal{L}^{(k)} = \mathbb{E}\left[\mathcal{L}\left(\theta^{(k)}_{100}\right)\right]$.

The algorithms evaluated are
SGD with fixed learning rate $\eta_0 \in [10^{-6}, 10]$,
SGD with annealing with decay factor in $[10^{-2},1]$ and initial rates $\eta_0$,
SGD with momentum (regular or Nesterov's variant~\cite{nesterov1994interior}) $[0.1, 0.999]$ and initial rates $\eta_0$,
SGD with parameter averaging~\cite{} with decay term in $[10^{-4}, 0.5]$ and exponent in $\{\frac{1}{2},\frac{3}{4}, 1\}$,
ADAGRAD~\cite{DuchiHS11} with initial rates $\eta_0$,
ADADELTA~\cite{zeiler2012adadelta} with decay parameter  $(1-\gamma) \in [10^{-4},0.5]$ and regularizer in $[10^{-6},10^{-2}$,
the incremental delta-bar-delta algorithm (IDBD~\cite{sutton1992adapting}),
RPROP~\cite{riedmiller1993direct} with initial stepsizes $\eta_0$,
RMSprop~\cite{tieleman2012lecture} with minimal learning rates $\eta_0$, maximal learning rates in $[10, 10^3]$ and decay parameter $\gamma$,
as well as conjugate gradients.
For the hyper-parameters ranges, we always consider one value per order of magnitude, and exhaustively sweep all combinations.

\subsection{Reference performance}
Each unit test is associated with a reference performance $\mathcal{L}_{sgd}$, and a corresponding reference learning rate $\eta_{best}$ that is determined
by doing a parameter sweep over all fixed learning rates for SGD (34 values log-uniform between $10^{-10}$ and $10$) and retaining the best-performing one.

In our aggregate plots, unit tests are sorted (per group) by their reference learning rate, i.e., those that require small steps on the left, and those where large steps are best on the right.
Algorithm setups are sorted as well, on the vertical axis, by their median performance on a reference unit test (quadratic, additive noise).

\subsection{Qualitative Evaluation}

The algorithm performance $\mathcal{L}^{(k)}$ is converted to a normalized value
$\mathcal{L}_{norm}^{(k)} = \frac{\mathcal{L}^{(k)} - \mathcal{L}_{init}}{\mathcal{L}_{sgd} - \mathcal{L}_{init}}$
where $\mathcal{L}_{init} = \mathbb{E}[\mathcal{L}(\theta_{0})]$ is the expected loss value at the initial point, similar to the approach taken in~\cite{Schaul2013}, but even more condensed.
In other words, a normalized value near zero corresponds to no progress, negative denotes divergence, and a value near one is equivalent to the best SGD.
Based on these results, we assign a qualitative color value to the performance of each algorithm setup on each unit test, to able to represent it in a single pixel in the resulting figures: 
\begin{itemize}
\item \textbf{Red:} Divergence or numerical instability in all run.
\item \textbf{Violet:} Divergence or numerical instability in at least one run.
\item \textbf{Orange:} Insufficient progress: $\text{median}(\mathcal{L}_{norm}) < 0.1$ 
\item \textbf{Yellow:} Good progress: $\text{median}(\mathcal{L}_{norm}) > 0.1$ and high variability: $\mathcal{L}_{norm} < 0.1$ for at least $\frac{1}{4}$ of the runs.
\item \textbf{Green:} Good progress: $\text{median}(\mathcal{L}_{norm}) > 0.1$ and low variability: $\mathcal{L}_{norm} < 0.1$ for at most $\frac{1}{4}$ of the runs.
\item \textbf{Blue:} Excellent progress: $\text{median}(\mathcal{L}_{norm}) > 2$.
\end{itemize}

\begin{sidewaysfigure}[hp]
%\begin{figure}[hp]
%\begin{center}
\hspace{-5em}
\includegraphics[width=1.15\columnwidth, angle =0 ]{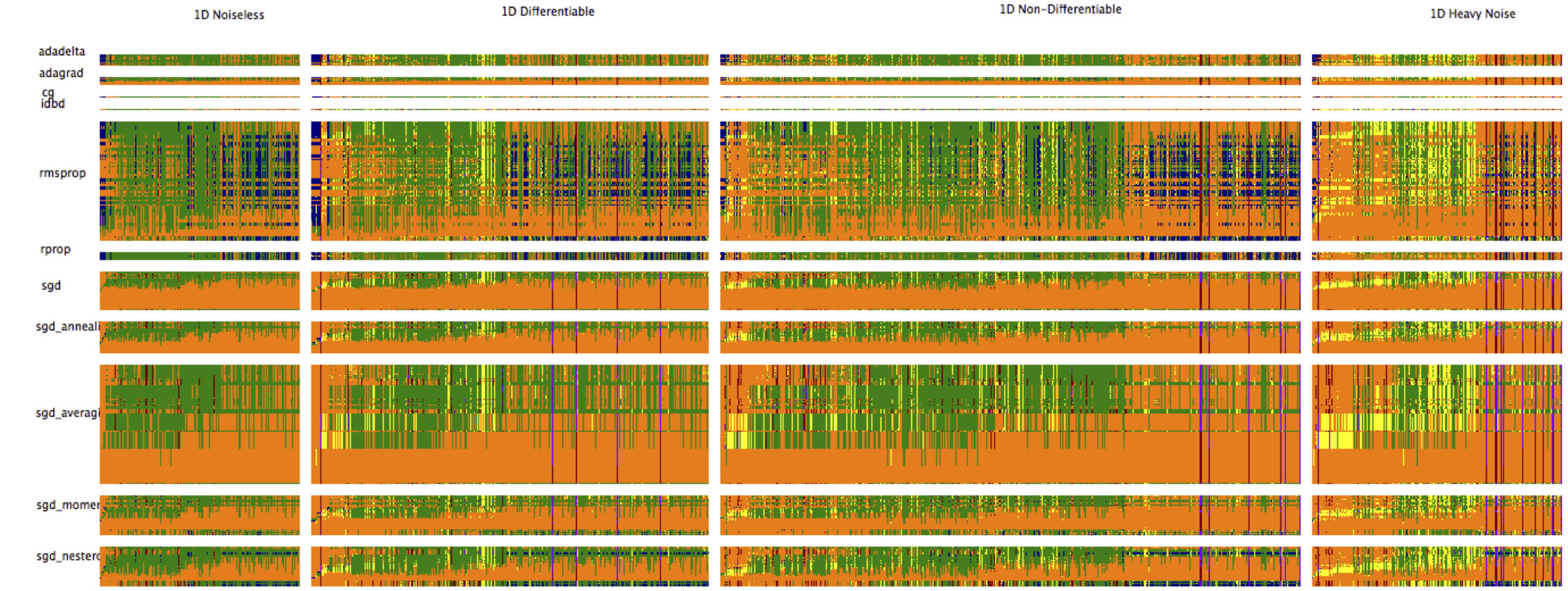}
\caption{\label{onedim}
Qualitative results for all algorithm variants (350) on all stationnary, one-dimensional unit tests.
Each column is one unit test, grouped by shared properties (see caption), for example the first group includes all noise-free 1D unit tests, where groups of unit tests can be partially overlapping. Each group of rows is one algorithm, with one set of hyper-parameters per row. The color code is: red/violet=divergence, orange=slow, yellow=variability, green=acceptable, blue=excellent (see main text for details).
}
%\end{center}
%\end{figure}
\end{sidewaysfigure}

\begin{sidewaysfigure}[hp]
%\begin{figure}[hp]
%\begin{center}
\hspace{-5em}
\includegraphics[width=1.15\columnwidth, angle =0 ]{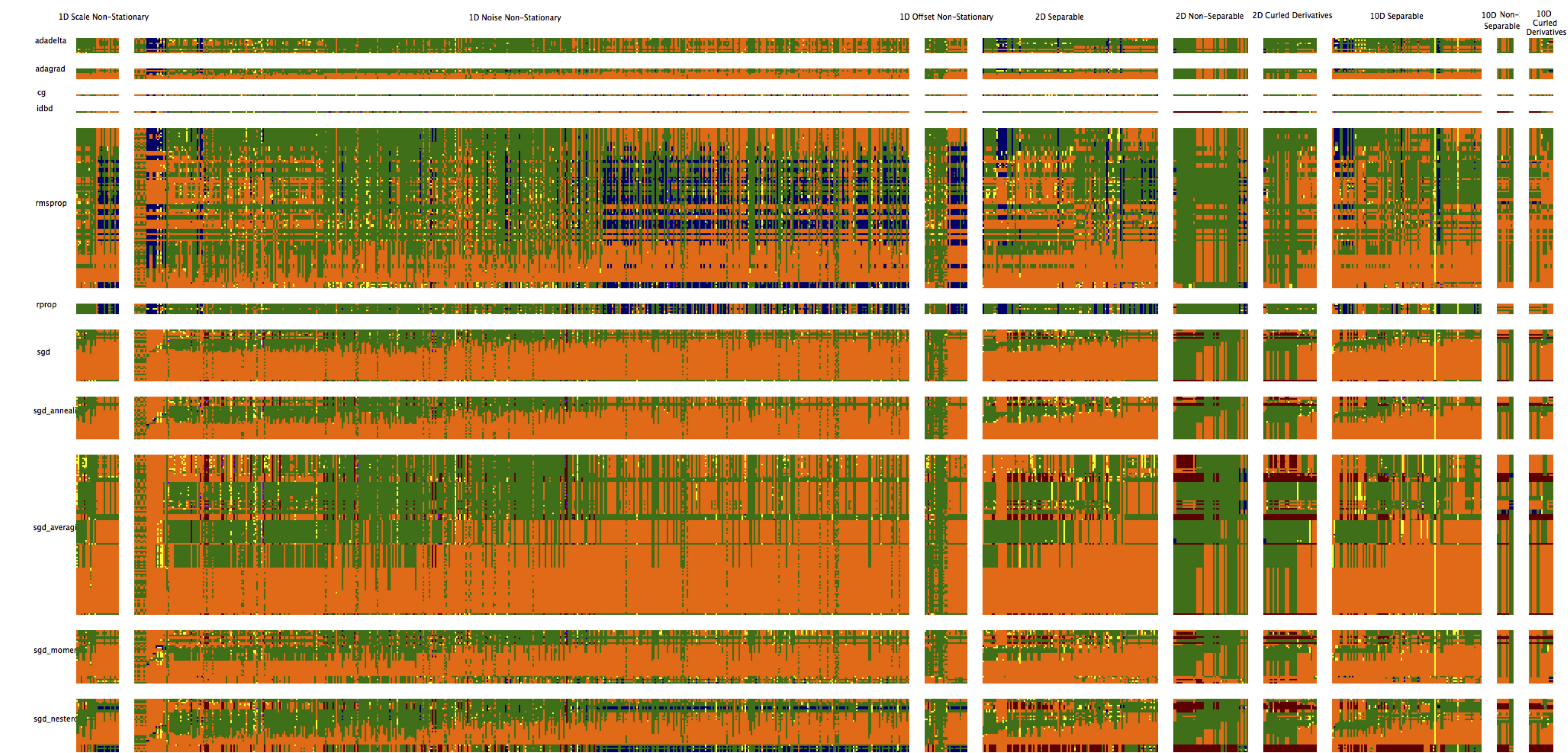}
\caption{\label{ndim}
Qualitative results for all algorithm variants (350) on all non-stationary one-dimensional unit tests (firs three groups), on all two-dimensional ones (next three groups), and on all ten-dimensional ones (last three groups).
The color code is: red/violet=divergence, orange=slow, yellow=variability, green=acceptable, blue=excellent. See Figure~\ref{onedim} and main text for details.
}
%\end{center}
%\end{figure}
\end{sidewaysfigure}

\subsection{Results}

Figures~\ref{onedim} and~\ref{ndim}  shows the qualitative results of all algorithm variants on all the unit tests. There is a wealth of information in these visualizations. For example the relatively scarce amount of blue indicate that it is difficult to substantially beat well-tuned SGD in performance on most unit tests.
Another unsurprising conclusion is that hyper-parameter tuning matters much less for the adaptive algorithms (ADAGRAD, ADADELTA, RPROP, RMSprop) than for the non-adaptive SGD variants. Also, while some unit tests are more tricky than others on average, there is quite some diversity in the sense that some algorithms may outdo SGD on a unit test where other algorithms fail (especially on the non-differentiable functions). 
%Of course there is also a vast amount of finer-grained observations on which shape prototypes are difficult for which algorithms for example, and what goes on in the diverging unit testes, etc. but space limitations keep us from doing this analysis here.

\section{Realism and Future Work}

We do not expect to replace real-world benchmark domains, but rather to complement them with our suite of unit tests. 
Still, it is important to have sufficient coverage of the types of potential difficulties encountered in realistic settings.
To a much lesser degree, we may not want to clutter the test suite with unit tests that measure issues which never occur in realistic problems.

\begin{figure}[tb]
\begin{center}
\includegraphics[width=0.49\columnwidth]{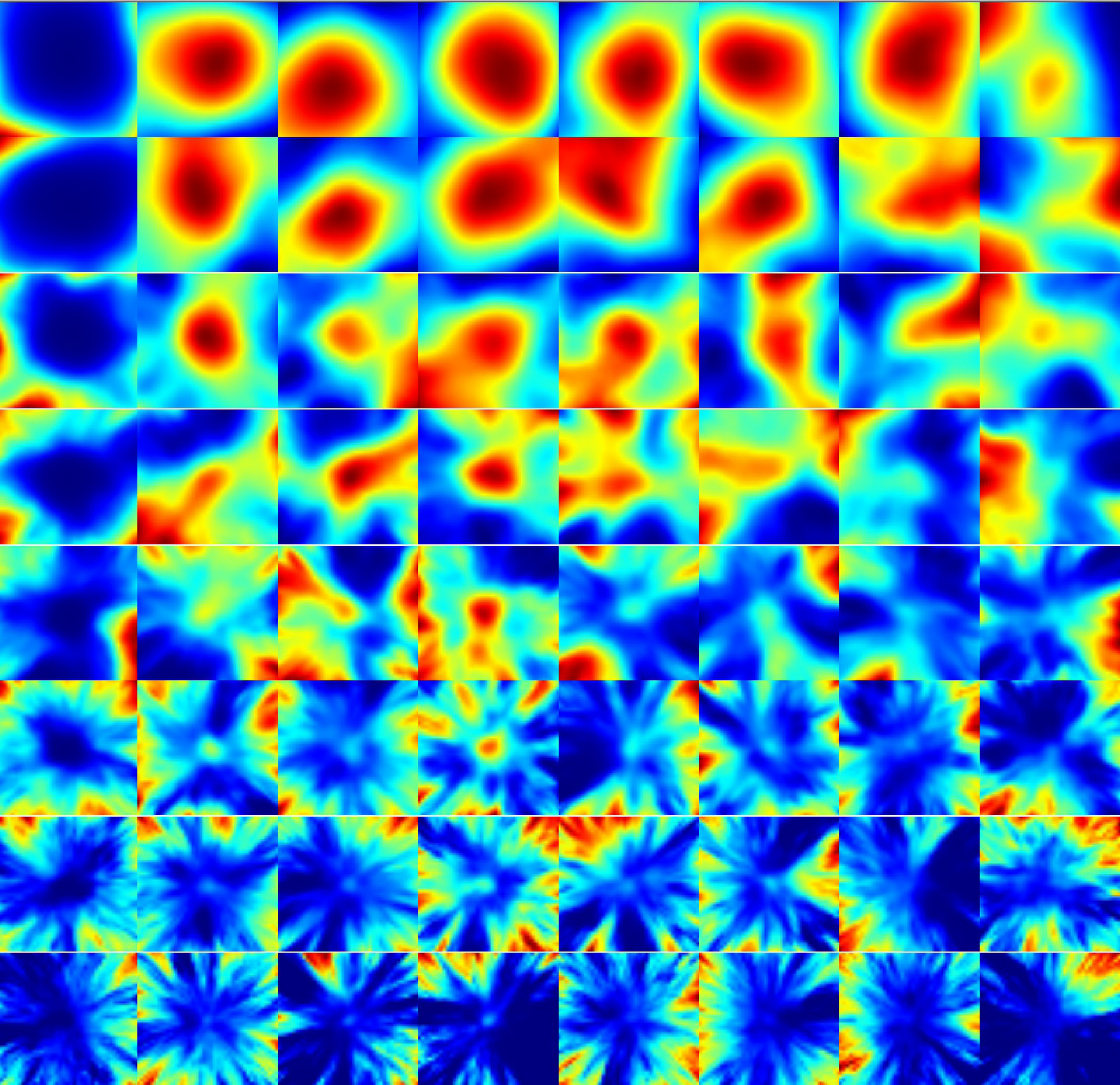}
\includegraphics[width=0.49\columnwidth]{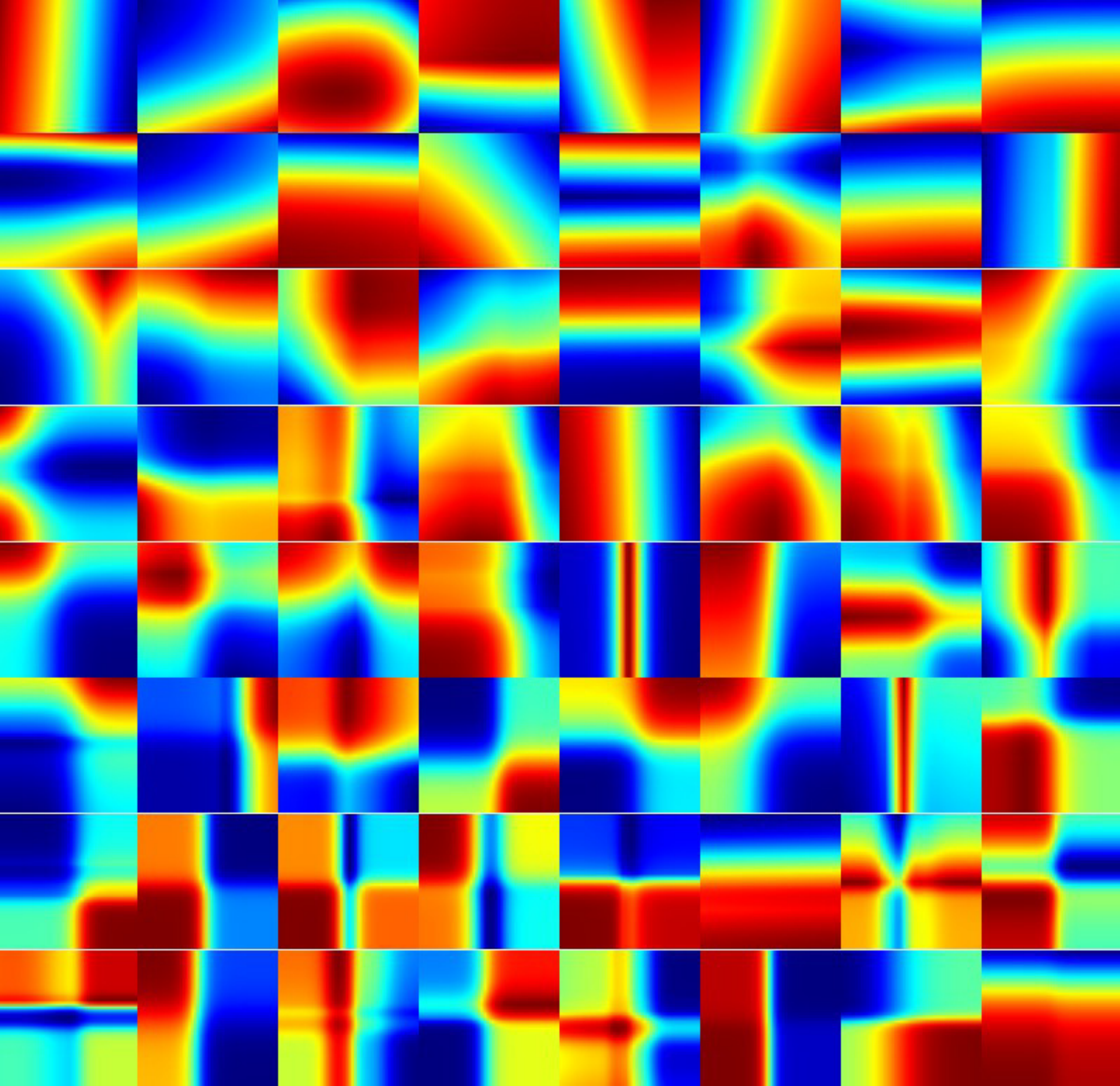}
\caption{\label{MnistProjections} \textbf{Left:} collection of 64 random projections into two dimensions of the MNIST loss surface (based on one randomly sampled digit for each column). The projections are centered around the weights learned after one epoch of training, and different projections are plotted on scales between 0.05 (top row) and 0.5 (bottom row). 
\textbf{Right:} the same as on the left, but with axis-aligned projections.}
\end{center}
\end{figure}

It is not straightforward to map very high-dimensional real-world loss functions down to low-dimensional prototype shapes, but it is not impossible. 
For example, in Figure~\ref{MnistProjections} we show some random projections in parameter space of the loss function in an MNIST classification task with an MLP~\cite{lecun-cortes-98}.
We defer a fuller investigation of this type, namely obtaining statistics on how commonly different prototypes are occurring, to future work.

\begin{figure}[tb]
\begin{center}
\includegraphics[width=\columnwidth]{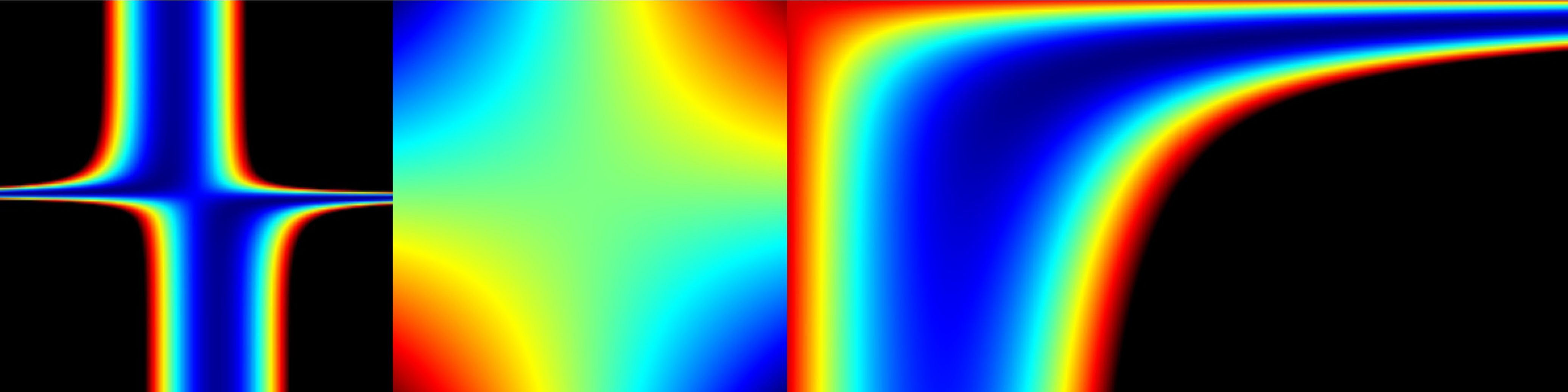}
\caption{\label{Aefig}
Illustration of the loss surface of a one-dimensional auto-encoder, as defined in the text, where the darkest blue corresponds to the lowest loss. \textbf{Left:} from the zoomed-out perspective if appears to be roughly a vertical valley, leading an optimizer toward the y-axis from almost anywhere in the space. \textbf{Center:} the zoomed-in perspective around the origin, which is looking like a prototypical saddle point. \textbf{Right:} the shape of the valley in the lower left quadrant, the walls of which become steeper the more the search progresses.}
\end{center}
\end{figure}

However, the unit tests capture the properties of some examples that \emph{can} be analyzed. One of them
was discussed in section~\ref{sec:curl}, another one is the simple loss function of a one-dimensional auto-encoder:
\[
\mathcal{L}_{\theta}(x) = \left(x + \theta_2 \cdot  \sigma(x \cdot \theta_1) \right)^2
\]
where $\sigma$ is the sigmoid function.
Even in the absence of noise, this minimal scenario has a saddle-point near $\theta= (0,0)$, a plateau shape away from the axes,
a cliff shape near the vertical axis,
and a correlated valley near $\theta=(1,1)$, as illustrated in Figure~\ref{Aefig}.
All of these prototypical shapes are included in our set of unit tests.

An alternative approach is predictive: if the performance on the unit tests is highly predictive of an algorithm's performance on a some real-world task, then those unit tests must be capturing the essential aspects of the task. Again, building such a predictor is an objective for future work.

%\subsection{[Optional] Non-stationarity and Algorithm Dynamics}
\subsection{Algorithm Dynamics}
\label{algostates}

Our long-term objective is to be able to do systematic testing and a full investigation of 
the optimization dynamics for a given algorithm.
Of course, it is not possible to test it exhaustively on all possible loss functions (because there are infinitely many),
but a divide-and-conquer approach may be the next best thing.
For this, we introduce the notion of \emph{algorithm state}, which is changing 
during optimization (e.g., the current stepsize or momentum).
Now, a long optimization process can be seen as the \emph{chaining} of 
a number of unit tests, while preserving the algorithm state in-between them.
Our hypothesis is that the set of all possible {chains} of unit tests in our collection covers most of the qualitatively different (stationary or non-stationary) loss functions an optimization algorithm may encounter.

%More formally, for each algorithm $a$, we consider the Markov decision process $(S_a, U, P, R)$, with algorithm states $s \in S_a$, unit tests from our collection $u \in U$ interpreted as actions, and transition dynamics that describe how the algorithm state changes while running on a specific unit test: $P_u(s' | s)$, where $s'$ is the final state after starting in $s$ and running $u$. Note that an algorithm's default initial state $s_0$ is only one choice out of many to state a unit test with. One possible reward model is the normalized loss progress $R(s, u) = f_{norm}$.

%\subsubsection{Adversarial Testing for Divergence}

To evaluate an algorithm's robustness (rather than its expected performance), we can assume that an \emph{adversary} picks the worst-case unit tests at each step in the sequence. An algorithm is only truly robust if it does not diverge under any sequence of unit tests.
%\subsubsection{Performance near Attractors}
Besides the worst-case, we may also want to study typical expected behavior, namely whether the dynamics have an \emph{attractor} in the algorithm's state space. 
If an attractor exists where the algorithm is stable, then it becomes useful to look at the secondary criterion for the algorithm, namely its expected (normalized) performance. 
We conjecture that this analysis may lead to novel insights into how to design robust and adaptive optimization algorithms.

%In that case, it may also be useful to initialize the algorithm such that $s_0 = s_*$.
%\subsubsection{State Discretization}
%For most algorithms, the space of all possible internal states is very large. In order to make the MDP-analysis tractable, we discretize the state space $S$. For example, in an algorithm where only the stepsize $\eta$ is adapted over time, we can use $s=\lfloor \log(\eta) \rfloor$. We then prune $S$ to only include those states reachable from the initial $s_0$ via running a chain of unit tests.

\section{Conclusion}

This paper established a large collection of simple comparative benchmarks to evaluate stochastic optimization algorithms, on a broad range of small-scale, isolated, and well-understood difficulties. 
This approach helps disentangle issues that tend to be confounded in real-world scenarios, while retaining realistic properties.
Our initial results on a dozen established algorithms (under a variety of different hyperparameter settings) show that robustness is non-trivial, and that different algorithms struggle on different unit tests.
The testing framework is open-source, extensible to new function classes, and easy to use for evaluating the robustness of new algorithms.

The full source code (see also Appendix A) is available under BSD license at:

\verb|        https://github.com/IoannisAntonoglou/optimBench|

\subsection*{Acknowledgements}
We thank the anonymous ICLR reviewers for their many constructive comments.

\bibliography{bib.bib}

\appendix
\section{Appendix: Framework Software}

As part of this work a software framework was developed for the computing and managing all the results obtained for all the different configurations of function prototypes and algorithms. The main component of the system is a database where all the results are stored and can be easily retrieved by querying the database accordingly. The building blocks of this database are the individual experiments, where each experiment is associated to a unit test and an algorithm with fixed parameters. An instance of an experiment database can either be loaded from the disk, or it can be created on the fly by running the associated experiments as needed. The code below creates a database and runs all the experiments for all the readily available algorithms and default unit tests, and then saves them to disk:  
\begin{verbatim}
 require 'experiment'  
 local db = experimentsDB()  
 db:runExperiments()  
 db:save('experimentsDB')  
\end{verbatim}

This database now can be loaded from the disk, and the user can query it in order to retrieve specific experiments, using \emph{filters}. An example is shown below:

\begin{verbatim}
 local db = experimentsDB()
 db:load('experimentsDB')
 local experiments = db:filter({fun={'quad', 'line'}, 
                                algo={'sgd'}, learningRate=1e-4})
\end{verbatim}

The code above loads an experiment database from the disk and it retrieves all the experiments for all the quadratic and line prototype shapes, for all different types of noise and all scales, further selecting the subset of experiments to those optimized using SGD with learningRate equal to 1e-4. The user can rerun the extracted experiments or have access to the associated results, i.e., the expected value of the function in different optimization steps, along with the associated parameters values. 
In order to qualitatively assess the results the following code can be used:

\begin{verbatim}
 db:ComputeReferenceValues()
 db:cleanDB()
 db:assessPerformance()
 db:plotExperiments()
\end{verbatim}

The code above computes the reference expected values for each prototype function, it removes the experiments for which no reference value is available, then it qualitatively assesses the performance of all the available experiments and finally it plots the results given the color configuration described in section 3.3. 
It is really easy to add a new algorithm in the database in order to evaluate its robustness. The code below illustrates a simple example:

\begin{verbatim}
 db:addAlgorithm(algoname, algofun, opt)
 db:testAlgorithm(algoname)
 db:plotExperiments({}, {algoname})
\end{verbatim}

Here a new algorithm with name \verb|algoname|, function instance \verb|algo| (which should satisfy the \verb|optim| interface), and a table of different parameter configurations \verb|opt| is added to the database and it is tested under all available functions prototypes. Finally, the last line plots a graph with all the results for this algorithm. 

It is also possible to add a set of new unit tests to the database, and subsequently run a set of experiments associated with them. There are different parameters to be defined for the creation of a set of unit tests (that allow wildcard specification too):
\begin{enumerate}
\item the concatenated shape prototypes for each dimension,
\item the noise prototype to be applied to each dimension,
\item the scale of each dimension of the function,
\item in case of multivariate unit tests, a parameter specifies which $p$-norm is used for the combination,
\item a rotation parameter that induces correlation of the different parameter dimensions, and
\item a curl parameter that changes the vector field of a multivariate function.
\end{enumerate}

\end{document}